\def\year{2021}\relax
\documentclass[letterpaper]{article} 
\usepackage{aaai21}  
\usepackage{times}  
\usepackage{helvet} 
\usepackage{courier}  
\usepackage[hyphens]{url}  
\usepackage{graphicx} 
\usepackage{natbib}  
\usepackage{caption} 
\usepackage{tabularx}
\usepackage{subcaption}

\usepackage{amsmath}
\usepackage[linesnumbered, ruled]{algorithm2e}
\usepackage{booktabs}
\usepackage{multirow}

\newtheorem{problem}{Problem}

\newtheorem{analysis}{Analysis}
\newtheorem{proof}{Proof}

\nocopyright
\title{Simultaneous Relevance and Diversity:\\A New Recommendation Inference Approach}

\iftrue
\author {
        Yifang Liu\textsuperscript{\rm 1},
        Zhentao Xu\textsuperscript{\rm 1},
        Qiyuan An\textsuperscript{\rm 2},
        Yang Yi\textsuperscript{\rm 2},
        Yanzhi Wang\textsuperscript{\rm 3},
        Trevor Hastie\textsuperscript{\rm 4} \\
}
\affiliations {
    \textsuperscript{\rm 1} Smule,
    \textsuperscript{\rm 2} Virginia Tech,
    \textsuperscript{\rm 3} Northeastern University,
    \textsuperscript{\rm 4} Stanford University \\
    ifnliu@gmail.com, frankxu@umich.edu, {anqiyuan,yangyi8}@vt.edu, yanz.wang@northeastern.edu, hastie@stanford.edu
}
\fi
\begin{document}

\maketitle

\begin{abstract}
Relevance and diversity are both important to the success of recommender systems, as they help users to discover from a large pool of items a compact set of candidates that are not only interesting but exploratory as well. The challenge is that relevance and diversity usually act as two competing objectives in conventional recommender systems, which necessities the classic trade-off between exploitation and exploration. Traditionally, higher diversity often means sacrifice on relevance and vice versa. We propose a new approach, heterogeneous inference, which extends the general collaborative filtering (CF) by introducing a new way of CF inference, negative-to-positive. Heterogeneous inference achieves divergent relevance, where relevance and diversity support each other as two collaborating objectives in one recommendation model, and where recommendation diversity is an inherent outcome of the relevance inference process. Benefiting from its succinctness and flexibility, our approach is applicable to a wide range of recommendation scenarios/use-cases at various sophistication levels. Our analysis and experiments on public datasets and real-world production data show that our approach outperforms existing methods on relevance and diversity simultaneously. 
\end{abstract}

\section{Introduction}
\label{sec:introduction}
Relevance and diversity are both important to recommender systems, as they help users to discover from a large pool of items a compact set of candidates that are not only interesting/relevant but inspiring/exploratory as well. 
\par
As a popular recommendation approach, the conventional collaborative filtering (CF) focuses primarily on inferring the relevance of items to individual users. The relevance is usually translated from inter-item similarity, which can be inferred by positive feedback to different items from the same user(s). Because both the source item and target item in the inter-item similarity inference are positive feedback (a.k.a. positive engagement), we consider conventional CF being based on positive-to-positive (\textbf{p2p}) inference. 
\par
CF by p2p is adopted in a wide variety of successful recommendation algorithms. 
A main strength of CF is the high relevance and precision of its recommendations, while it also leads to a challenge for CF --- shrinking recommendation diversity: over time, the recommendation becomes more and more focused on what the user has shown strong interest in, while the scope of predicted user interest becomes narrower and narrower. As a result, less variety of content is recommended to the user. The converging recommendation scope clearly impairs the user's exploration horizon. In the long term, the converging recommendation feedback loop excludes a large space of potentially interesting items/topics from recommendations to the user. For this reason, this paper considers conventional CF as convergent relevance (\textbf{CR}) oriented. In other words, on the trade-off between exploitation vs. exploration, conventional CF tends to weigh mostly on exploitation at the cost of poor exploration. The concern of shrinking diversity is also known as the Rabbit Hole problem, the Filter Bubble issue, or the Echo Chamber effect in different domains~\cite{jiang2019degenerate,ge2020understanding,antikacioglu2017post,nguyen2014exploring,knijnenburg2016recommender}. 
\par
To tackle this challenge at its core, we propose a new recommendation approach, \textbf{Heterogeneous Inference (HI)}, which fundamentally extends the CF approach by introducing into CF a new channel of relevance inference, negative-to-positive (\textbf{n2p}) inference, in addition to the existing \textbf{p2p} inference. Like p2p, n2p makes the observation that when users are not interested in one item (i.e., the negative) they tend to be interested in some other items (i.e., the positive). 
In a loose statistical sense, "similarity" in relevance inference context can be viewed as a proxy for "correlation" between entities. CF only cares about positive correlation (p2p). HI leverages both positive correlation (p2p) and negative correlation (n2p) in one cohesive model.
\par
HI's potential to address the shrinking diversity problem can be mostly attributed to the divergent nature of n2p inference. Intuitively, there are many possible positives given one negative. Therefore, n2p is able to infuse diversity into its own relevance inference. By n2p inference, both relevance and diversity are inherent outcomes of the same relevance inference process. The intrinsically-diverse relevance is at the core of our \textbf{Divergent Relevance (DR)} concept. 
\par
\textbf{The main contribution} of this paper includes:
\par
(1) We present a new concept, divergent relevance (DR), for achieving relevance and diversity collaboratively in recommendations as two inherent outcomes of one relevance inference process. This is realized by our new recommendation approach, heterogeneous inference (HI), which extends CF with a negative-to-positive (n2p) inference component. 
\par
(2) HI provides a flexible, general framework for divergent relevance inference, which is suitable for different levels of model/algorithm sophistication and is applicable to a variety of scenarios in recommendation. HI can work either as a standalone recommendation algorithm (e.g., a real-time personalized recommender, a candidate generator) or as a module integrated in a sophisticated recommendation model (e.g., an embedding module inside a neural network). 
\par
(3) HI's effectiveness and efficiency are demonstrated through evaluations on both a well-known public recommendation dataset and real-world industrial production data.

\section{Preliminaries}
\label{sec:prelim}

\begin{figure*}[t] 
\centering
\includegraphics[width=5.5in,angle=0]{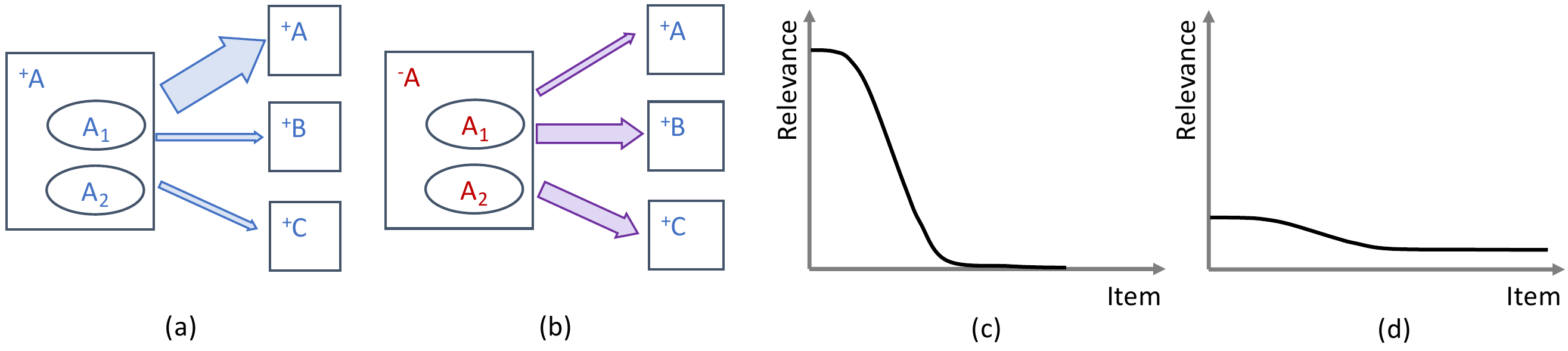}
\caption{Distribution of relevance over items. (a) p2p convergent relevance: precise and narrow relevance prediction. (b) n2p divergent relevance: relevance prediction spread out across all topics. (c) p2p relevance distribution: largely skewed. (d) n2p relevance distribution: relatively flat and even.}
\label{fig:diverge-distr}
\end{figure*}

In general, CF and HI are applicable to both implicit feedback (e.g., user selecting/skipping an item) and explicit feedback (e.g., user rating an item).
User feedback can assume different value types, e.g., real values, discrete numbers, and categories. For the sake of clear demonstration and comparison, this paper explains HI algorithms with binary user feedback observations. Any real-valued feedback is converted into either a positive feedback or a negative feedback according to a preset threshold, before being used as the label of a data point during evaluation.

\subsection{Notation and problem statement} 
Given an set of observed user-item engagement examples,

$\mathbf{X}$ \mbox{\ \ \ \ \ } An m-by-n matrix representing the positive feedback of $m$ users to $n$ items. Each element $x_{i,j} \in \{0,1\}$ indicates whether user $i$ has a positive feedback to an impression of item $j$: $1$ means positive feedback, $0$ means negative or unobserved feedback. 

$\mathbf{O}$ \mbox{\ \ \ \ \ } An m-by-n matrix representing the observation of $m$ users' feedback to $n$ items. Each element $o_{i,j} \in \{0,1\}$ indicates whether user $i$ has an impression on item $j$: $1$ means observed, $0$ means not observed. Note that $\mathbf{O}$ is the information usually ignored by conventional recommendation algorithms. 

$\Tilde{x}_{i,j}$ \mbox{\ \ \ } An unobserved data point (i.e., $o_{i,j}=0$), whose binary value indicates whether user $i$ has a positive feedback to item $j$, when item $j$ is presented to user $i$. 

\begin{problem}
\label{def:problem}
\textbf{Engagement Prediction}
Given a set of observed engagements by a group of users with a group of content items, predict $p(\Tilde{x})$, the likelihood of a given user giving positive feedback to a specific item when the corresponding impression presents. In other words, the problem is to estimate the following posterior:  
\begin{equation}
\begin{aligned}
\label{equ:problem}
    p(\Tilde{x} | \mathbf{X}, \mathbf{O})
\end{aligned}     
\end{equation}
where $\Tilde{x}$ can be any hypothetical future impression of item j on user i, $\Tilde{x}_{i,j}$. 
\end{problem}
\par
In conventional CF recommenders, because the information in $\mathbf{O}$ is usually overlooked, the engagement prediction target degenerates to a simpler posterior: 
\begin{equation}
\begin{aligned}
\label{equ:conventional_problem}
    p(\Tilde{x} | \mathbf{X})
\end{aligned}    
\end{equation}

\subsection{Related work}
If the relevance of item $j$ to user $i$ is implied by the likelihood of the user giving positive feedback to (being interested in) the item, i.e., $p(x_{i,j} = 1)$, recommendation diversity can be interpreted as the distribution of the relevance over different recommendation items. The wider the relevance is spread out across different items, the higher the diversity is. 
\par
\underline{Conventional CF}.
CF comes in various flavors of algorithms or models. For example, user/item collaborative filtering~\cite{www15zhao,recsys10davidson,nips08mnih,icml08salak,nips05srebro,ikdm14facebook,tmis16gomez}, deep learning~\cite{recsys16youtube,arxiv18krichene,recsys16cheng,icml07hinton,nips13oord,kdd15wang,www17pinterest}, deep embedding models~\cite{arxiv18krichene,kdd17okura}, Factorization Machines~\cite{icdm10rendle}, Matrix Factorization~\cite{computer2009koren, www15zhao,nature99lee,nips08mnih,icml08salak,nips05srebro}, ALS~\cite{jlmr15hastie}, SVD++~\cite{kdd08koren}, PITF~\cite{wsdm10rendle}, and FPMC~\cite{www10rendle}.
\par
Regardless of the specific algorithm, recommendation relevance in CF relies on certain forms of p2p inference, which is derived from the user-item engagement matrix, $\mathbf{X}$. Recommendation diversity is the objective of a separate model/algorithm component (other than CF), which uses various different strategies explained later in this section. In existing recommendation approaches, diverse recommendation means deviation from top-relevance recommendation, i.e., diversity is in the opposite direction of conventional CF. Therefore,  in traditional recommender systems the CF component and the diversity component compete against each other in two separate processes. Conventional CF has to trade a certain degree of relevance for a certain level of diversity in the recommendations, i.e., make a trade-off between the two competing objectives.  
\par
A batch recommendation process usually works in two stages: candidate generation and ranking.
Next, we explain how CF by p2p works in matrix factorization fashion for candidate generation.
\par
\mbox{Step 1. } \mbox{rank-$k$ matrix factorization of $\mathbf{X}^T \mathbf{X}$:}
\begin{equation}
\begin{aligned}
\mathbf{X}^T \mathbf{X} \approx \mbox{\ } \mathbf{P} \mbox{\ } \mathbf{Q}, \mbox{\ where\ } \mathbf{P} \in \Re^{n\times k}, \mathbf{Q} \in \Re^{k\times n}     \label{equ:pq}
\end{aligned}
\end{equation}

\mbox{Step 2. } \mbox{smooth approximation of $\mathbf{X}^T \mathbf{X}$:}
\begin{equation}
\begin{aligned}
\mathbf{C} = \mbox{\ } \mathbf{P} \mbox{\ } \mathbf{Q}, \mbox{\ where\ } \mathbf{C} \in \Re^{n\times n}      \label{equ:c}
\end{aligned}
\end{equation}

\mbox{Step 3. } \mbox{ inferred items' relevance scores per user:}
\begin{equation}
\begin{aligned}
^*\mathbf{X} = \mbox{\ } \mathbf{X} \mbox{\ } \mathbf{C}, \mbox{\ where\ } ^*\mathbf{X} \in \Re^{m\times n}    \label{equ:x}
\end{aligned}
\end{equation}
Step 1 obtains the input matrix of p2p item-to-item similarity, $\mathbf{X}^T \mathbf{X}$, by multiplying the transpose of the user-item positive engagement matrix with itself. It indicates the connection between items that received positive feedback from the same users. 
Step 2 calculates $\mathbf{C}=\mathbf{P} \mathbf{Q}$, a smoothed approximation of the p2p inter-item similarity matrix, based on the rank-$k$ matrix factorization from Step 1. $\mathbf{C}$ replaces zeros in $\mathbf{X}^T \mathbf{X}$ with similarity values estimated by the factorization. 
Step 3 estimates the relevance scores of all items for every individual user, by multiplying the original user-item matrix with the smoothed p2p inter-item similarity matrix. The recommendation candidates for each user can be selected as the items with the highest relevance scores in the corresponding row of matrix $^*\mathbf{X}$. This candidate relevance estimation procedure is typical CF by p2p. 
\par
Figure~\ref{fig:diverge-distr} (a) explains the convergent relevance (CR) nature of the CF by p2p,
which causes the issue of shrinking recommendation diversity. Suppose we have 3 top-level latent topics: $A$, $B$, and $C$. Under each of these topics, we have sub-topics: 1, 2, and 3. Items similar to those in $A_1$ and $A_2$ that received positive feedback, will most likely fall under the same top-level topic, $A$. Most of them may even belong to one of the sub-topic there, e.g., $A_1$, which may be the focus of the next group of recommendations. Over time, the items recommended through p2p inference may converge to a highly-relevant but narrowing topic. Figure~\ref{fig:diverge-distr} (c) illustrates the skewed distribution of p2p relevance across all items.    
\par
\underline{Recommendation diversity improvement}.
A large amount of research has been devoted into enhancing recommendation diversity. The most direct method of enhancing diversity is using calculated metrics (e.g. Entropy \cite{infosci17Tommaso}, Item Similarity Scores \cite{bradley2001improving,castagnos2013diversity,l2014understanding}) as proxies of diversity (e.g. Item Popularity \cite{acm11hurley,ieee12Ado}) to augment the scoring models of existing recommender systems. Some applications leverage content-specific data, such as tags and topics \cite{kbs16Zhang,vargas2014coverage}, to ensure a good mix of items across different categories, i.e., to improve recommendation's genre coverage. Calibrated Recommendations~\cite{recsys18steck} propose a way to encourage the recommendations in different categories/areas, based on the user's interest distribution. More sophisticated approaches include graph-based algorithms that formulate diversity as a max-flow problem \cite{divers11Adomavicius} or Markov Chain \cite{acm16paudel}. Matrix factorization methods characterize item similarities by projecting their properties as latent features \cite{recsys17paudel}.  
\par
Most of existing methods treat diversity as a competing objective against relevance. They improve recommendation diversity in a separate process (usually as a re-ranking criterion) from the one used for recommendation relevance improvement. As a result, any increase in recommendation diversity would hurt the quality of recommendation relevance, under the assumption that one has to choose between diversity and relevance in the trade-off.
\par
\underline{Negative feedback in deep learning}. 
Recently, some research is done to leverage both positive and negative feedback inside a deep learning model for recommendations, e.g., deep reinforcement learning~\cite{kdd18zhao,zou2019reinforcement}, deep neural network for interaction embedding~\cite{xie2014ijcai,zhang2019deep,gauci2018horizon}. These methods achieved notable improvement on recommendation relevance, while their impact on recommendation diversity is unknown. 
\par
These methods do not provide a general, self-contained framework for relevance inference, i.e., the use of negative feedback is mixed into the overall recommendation model on a case-by-case basis. In fact, their implicit inference with negative feedback acts in an accessory role to support the main recommendation model for its use case. Meanwhile, their model complexity due to the tailored DNN design makes them impractical for time-constrained use cases, e.g., realtime personalized recommendation.    

\section{Our approach}
\label{sec:dr}
This section starts with establishing the crucial foundation of HI on our general-purpose feedback encoding scheme. Then, we present how HI simultaneously achieves relevance and diversity as two collaborating objectives in the inherent outcomes of one relevance inference model. Finally, exemplary algorithm implementations of HI for several recommendation scenarios/use-cases are explained.  

\subsection{Feedback-Differentiating Encoding}
Conventional feedback encoding (CONFE), $\mathbf{X}$, does not differentiate unknown feedback from negative feedback --- usually, they are represented by the same values or are both missing in the dataset.   
\par
One straightforward way to encode the difference between negative and unknown is to represent negative with a number (e.g., $0$ or $-1$) and unknown as a non-number (e.g., Null or NaN) in the implementation of the CF algorithm. For example, Apache Spark realizes this type of encoding in their ALS algorithm implementation. However, this negative-unknown differentiating capability is not always supported in CF algorithm implementations.
\par
Therefore, we propose a general-purpose feedback-differentiating encoding (FEEDE) scheme, which precisely encodes different types of feedback, regardless whether the CF algorithm implementation supports it or not. FEEDE demonstrates the feedback-differentiating capability by complementing the conventional positive feedback encoding, $\mathbf{X}$, with an additional negative feedback encoding, $\mathbf{Y}$,
\begin{equation}
\begin{aligned}
\mathbf{Y} = \mathbf{O} - \mathbf{X}
\end{aligned}
\end{equation}
By this definition, in binary feedback context, element $y_{i,j}$ in $\mathbf{Y}$ represents whether an impression gets negative feedback: 
\begin{equation}
\begin{aligned}
\label{equ:encode_y}
    y_{i,j} =
    \begin{cases}
    1 & \mbox{\ \ \ $x_{i,j} = 0$ and $o_{i,j} = 1$}  \nonumber\\
    0 & \mbox{\ \ \ ($x_{i,j} = 1$ and $o_{i,j} = 1$) or $o_{i,j} = 0$}  \nonumber\\
    \end{cases}
\end{aligned}
\end{equation}
Note that for implicit feedback, FEEDE relies on an assumption that the impression of an item on a user is observable, i.e., $\mathbf{O}$ is known. FEEDE's advantage over CONFE is explained in the following analysis. 
\par
If the feedback is a range of integers (like typical ratings on a scale of $1$ to $h$), the definition of $\mathbf{Y}$ still works with an extra step of rating normalization. For example, with $\{1, ..., 5\}$ ratings on individual items:
each element $x_{i,j}$ in $\mathbf{X}$ is obtained as $x_{i,j} = Rating_{i,j} / 6$. Then, the same definition, $\mathbf{Y}=\mathbf{O}-\mathbf{X}$, implies:
\begin{equation}
\begin{aligned}
y_{i,j} =
    \begin{cases}
    1-x_{i,j} & \mbox{\ \ \ $o_{i,j} = 1$}  \nonumber\\
    0 & \mbox{\ \ \ $o_{i,j} = 0$}  \nonumber\\
    \end{cases}
\end{aligned}
\end{equation}

\begin{analysis}
\label{theory:feede}
For a engagement prediction problem, FEEDE feedback encodes more information in the predictive model than CONFE. The information gain of FEEDE over CONFE is given by the conditional mutual information, $I(\Tilde{x}; \mathbf{O} | \mathbf{X})$, or equivalently $I(\Tilde{x}; \mathbf{Y} | \mathbf{X})$.
\end{analysis}
\begin{proof}
Engagement prediction by CONFE in conventional recommender systems is specified by the conditional probability in Equation~\ref{equ:conventional_problem}. Engagement prediction by FEEDE in heterogeneous inference recommender systems is specified by the conditional probability in Equation~\ref{equ:problem}.  
\par
By the definition of conditional mutual information, we have
\begin{equation}
\begin{aligned}
& I(\Tilde{x}; \mathbf{O} | \mathbf{X}) = H(\Tilde{x} | \mathbf{X}) - H(\Tilde{x} | \mathbf{O}, \mathbf{X}) =  \nonumber\\
& \sum_{X \in \mathcal{X}} p(X) \sum_{\Tilde{x} \in \{0,1\}} \sum_{O \in \mathcal{O}} p(\Tilde{x}; O | X) \log \frac{p(\Tilde{x}; O | X)}{p(\Tilde{x} | X) p(O | X)} \nonumber
\end{aligned}
\end{equation}
where $H(\Tilde{x} | \mathbf{X})$ is the conditional entropy representing the amount of uncertainty in $\Tilde{x}$ given the observation of $\mathbf{X}$; $H(\Tilde{x} | \mathbf{O}, \mathbf{X})$ is the conditional entropy representing the amount of uncertainty in $\Tilde{x}$ given the observation of $\mathbf{X}$ and $\mathbf{O}$; $\mathcal{X}$ and $\mathcal{O}$ are the alphabet (the set of all possible values) of $\mathbf{X}$ and $\mathbf{O}$, respectively. 
\par
Since $H(\Tilde{x} | \mathbf{X})$ and $H(\Tilde{x} | \mathbf{O}, \mathbf{X})$ represent the entropy in the predictive conditional probability encoded by CONFE and FEEDE respectively, $I(\Tilde{x}; \mathbf{O} | \mathbf{X}) = H(\Tilde{x} | \mathbf{X}) - H(\Tilde{x} | \mathbf{O}, \mathbf{X})$ gives the information difference between them. $I(\Tilde{x}; \mathbf{O} | \mathbf{X}) > 0$ when random variables $\Tilde{x}$ and $O$ are not independent from each other given $X$, which is usually true in our feedback-based engagement prediction problem.
\par
By $\mathbf{Y}=\mathbf{O}-\mathbf{X}$, given either $(\mathbf{O}, \mathbf{X})$ or $(\mathbf{Y}, \mathbf{X})$, the other can be precisely calculated. Thus, they contain equivalent information --- $H(\Tilde{x} | \mathbf{O}, \mathbf{X})$ and $H(\Tilde{x} | \mathbf{Y}, \mathbf{X})$ have the same amount of uncertainty. Therefore,  
$ I(\Tilde{x}; \mathbf{O} | \mathbf{X}) = H(\Tilde{x} | \mathbf{X}) - H(\Tilde{x} | \mathbf{O}, \mathbf{X}) = H(\Tilde{x} | \mathbf{X}) - H(\Tilde{x} | \mathbf{Y}, \mathbf{X})
= I(\Tilde{x}; \mathbf{Y} | \mathbf{X}) $.
\end{proof}

\subsection{HI for divergent relevance: two goals, one model}
HI combines p2p and n2p inference in one cohesive recommendation model. It is able to gain relevance and diversity collaboratively as inherent outcomes of one relevance inference process, i.e., divergent relevance (DR). 

\textbf{HI inference framework for divergent relevance}. HI is built upon latent representations of entities (i.e., items and users), for example, via matrix factorization or embedding sub-network in a deep neural network. This way, the divergent relevance between two entities can be captured as the "similarity" between their latent representations. The latent representations are optimized/trained, according to the data of (positive and/or negative) correlation between two entities in the corresponding application/context. 

The divergent nature of n2p, i.e., relevance spreading over a wide variety of items in different topics, is illustrated in Figure~\ref{fig:diverge-distr} (b). Suppose that we have the same top-level latent topics and sub-topics as in the example from the previous section. Two items in $A_1$ and $A_2$, which received negative feedback from the user in the past, may suggest potential positive feedback to items in different top-level topics: $A$, $B$, and $C$ (mainly spread over the latter two). 
It is not hard to imagine that over time the recommendation based on n2p inference is not likely to converge to a narrowing topic. This is also reflected in the relatively flat distribution of n2p relevance over all items, illustrated by Figure~\ref{fig:diverge-distr} (d).
\par
From the algorithm perspective, HI is an extension of CF approach. Because of its succinctness and generality, HI can work as a standalone recommender algorithm, or it can be integrated as an embedding module into a sophisticated recommendation model. Next, we demonstrate several exemplary algorithms, which implement HI for three different recommendation scenarios: candidate generation, ranking, and realtime personalized recommendation.  

\begin{figure*}[t]
\centering
\includegraphics[width=5in,angle=0]{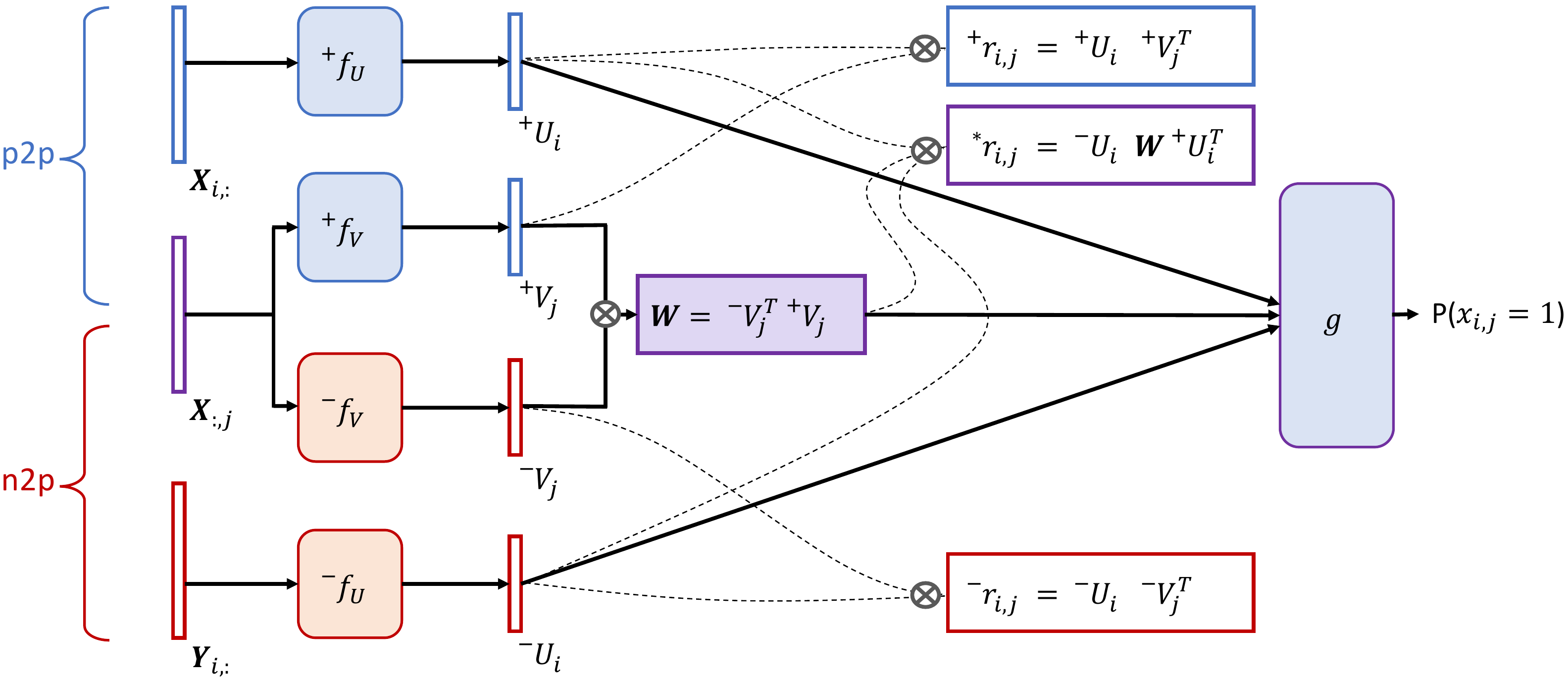}
\caption{Model architecture. Input vectors pass through embedding modules ($^+f_U$, $^+f_V$ for p2p inference; $^-f_U$, $^-f_V$ for n2p inference; $\mathbf{W}$ for p2p-n2p interaction). Then, the latent representation goes through $g$ to output. First, embedding modules are trained and fixed. Then, $g$ is trained.}
\label{fig:hi-architecture}
\end{figure*}

\subsection{Candidate generation by HI}
HI candidate generation is a combination of p2p and n2p similarity estimation. The two channels of inference (p2p and n2p) can be mixed by a preset or a dynamic candidate ratio, e.g., p2p vs. n2p = 70\% vs. 30\%.  Next, we outline candidate generation by n2p in matrix factorization style.
\par
\mbox{Step 1. } \mbox{rank-$k$ matrix factorization of $\mathbf{Y}^T \mathbf{X}$: \ \ \ }
\begin{equation}
\begin{aligned}
\mathbf{Y}^T \mathbf{X} \approx \mbox{\ } \mathbf{R} \mbox{\ } \mathbf{S}, \mbox{\ where\ } \mathbf{R} \in \Re^{n\times k}, \mathbf{S} \in \Re^{k\times n} \label{equ:rs}
\end{aligned}
\end{equation}

\mbox{Step 2. } \mbox{smooth approximation of $\mathbf{Y}^T \mathbf{X}$: \ \ \ }
\begin{equation}
\begin{aligned}
\mathbf{D} = \mbox{\ } \mathbf{R} \mbox{\ } \mathbf{S},  \mbox{\ where\ } \mathbf{D} \in \Re^{n\times n}  \label{equ:d}
\end{aligned}
\end{equation}

\mbox{Step 3. } \mbox{inferred divergent candidate items per user: \ \ \ }
\begin{equation}
\begin{aligned}
^*\mathbf{Z} = \mbox{\ } \mathbf{Y} \mbox{\ } \mathbf{D}, \mbox{\ where\ } ^*\mathbf{Z} \in \Re^{m\times n} \label{equ:z}
\end{aligned}
\end{equation}
The three steps for n2p candidate generation are similar to those of the p2p candidate generation, while the key difference lies in that the original item-item similarity matrix is obtained by multiplying the transpose of the user-item negative engagement matrix with the user-item positive engagement matrix. As a result, the smoothed n2p item similarity matrix, $\mathbf{D}$, implies the correlation between positively-engaged items and negatively-engaged items by the same users. Thus, we call this relevance inference negative-to-positive (n2p) inference. 

\subsection{Ranking by HI}
Similar to candidate generation, HI ranking also integrates convergent relevance by p2p and divergent relevance by n2p. This can be realized by integrating p2p and n2p as embedding sub-models into an overall relevance prediction model. Because of DNN's strength in joint optimization of embedding and predicting sub-models, we choose it as the overall relevance prediction model to demonstrate HI ranking. 
\par
The DNN model architecture is illustrated in Figure~\ref{fig:hi-architecture}.
The upper left sub-network corresponds to the p2p inference channel. Specifically, the input of positive feedback user vector, $\mathbf{X}_{i,:}$, and item vector, $\mathbf{X}_{:,j}$, are passed through $^+f_U$ and $^+f_V$, and are mapped to their $k$-factor latent row vectors (in p2p context) $^+U_i$ and $^+V_j$, respectively. Here, $^+U_i$ and $^+V_j$ represent the latent factors of user $i$ and item $j$, respectively, when only positive feedback is used to infer the likelihood of positive feedback involving user $i$ and item $j$, respectively. 
\par
The lower left sub-network corresponds to the n2p inference channel. Row vectors $^-U_i$ and $^-V_j$ are the latent representations (in n2p context) of $\mathbf{Y}_{i,:}$ and $\mathbf{X}_{:,j}$, respectively. Here, $^-U_i$ and $^-V_j$ represent the latent factors of user $i$ and item $j$, respectively, when only negative feedback is used to infer the likelihood of positive feedback involving user $i$ and item $j$, respectively.
$^+r_{i,j}$ and $^-r_{i,j}$ are the relevance scores derived from p2p channel and n2p channel, respectively. 
\par
Matrix $\mathbf{W} \in \Re^{k \times k}$ represents the interaction between p2p and n2p latent factors. The interaction indicates how the likelihood of positive engagement by user $i$ on item $j$ is impacted by the positive feedback factors jointly with the negative feedback factors. Specifically, an element $w_{p,q}$ in interaction matrix $\mathbf{W}$ for item $j$ indicates the likelihood of a user likes item $j$, given that the user does not like latent topic $p$ but likes latent topic $q$. 
\par
The advantage of the multi-module architecture is three-fold: (1) Separate p2p and n2p modules help balance relevance and diversity. It can also speed up the training by using pre-trained sub-models. (2) Available approximations of the embedding modules help cold start and responsiveness at prediction time. (3) Individual modules can be selected or deselected to suit different purposes/objectives. 
\par
Training has two consecutive phases: (1) Backward propagation is performed to optimize the sub-networks of $^+f_U$, $^-f_U$, $^+f_V$, and $^-f_V$. They are trained using $x_{i,j}$ (representing user $i$'s feedback to item $j$ when $o_{i,j}=1$) as the identical common target for $^+r_{i,j}$, $^-r_{i,j}$ and $^*r_{i,j}$, which are feedback estimated by positive feedback, negative feedback, and positive-joint-negative feedback, respectively. Once the sub-networks between the input and $g$ are trained in phase 1, they are fixed. (2) Backward propagation optimizes the sub-network of $g$ (a number of dense layers), where the ground truth for $\hat{p}(\Tilde{x}_{i,j})$ is the value of $x_{i,j}$ (only if $o_{i,j}=1$). 
\par
\textbf{Training loss} of the entire embedding sub-network is
\begin{equation}
\begin{aligned}
\mathcal{L} = & (^+\mathcal{L}) + \alpha (^-\mathcal{L}) + \gamma (^*\mathcal{L}) + \mbox{\ } \\
& ^+\lambda (\sum_{i}\| ^+U_i \|_2^2 + \sum_{j}\| ^+V_j \|_2^2) + \\
& ^-\lambda (\sum_{i}\| ^-U_i \|_2^2 +  \sum_{j}\| ^-V_j \|_2^2) 
\label{equ:lossfunc}
\end{aligned}
\end{equation}
where
\begin{equation}
\begin{aligned}
^+\mathcal{L} = & \sum_{i,j} (x_{i,j} - (^+\mathbf{U}_i)\mbox{\ }(^+\mathbf{V}_j)^T)^2, \mbox{\ \ p2p loss} \nonumber\\
^-\mathcal{L} = & \sum_{i,j} (x_{i,j} - (^-\mathbf{U}_i)\mbox{\ }(^-\mathbf{V}_j)^T)^2, \mbox{\ \ n2p loss} \nonumber\\
^*\mathcal{L} = & \sum_{i,j} (x_{i,j} - (^-\mathbf{U}_i) \mathbf{W} (^+\mathbf{U}_i)^T)^2, \mbox{\ \ interaction loss} \nonumber
\end{aligned}
\end{equation}
Prediction operates in two scenarios: (1) Warm start: the user and the item both have enough engagement data for an established profile. Normal forward pass is made through the whole network.
(2) Cold start: the user does not have enough engagement data for an established profile. In order to obtain a reasonable prediction in near realtime, the latent vectors, ($^+U_i$, $^-U_i$, and $\mathbf{W}$), are approximated with pre-trained models. The forward pass starts from them and goes through $g$ to the output. 
The following approximation is based on the matrix factorization in candidate generation.
\begin{align}
    ^+U_i \approx \mbox{\ } & (\mathbf{XP})_{i,:}   \label{equ:approx+u}\\
    ^-U_i \approx \mbox{\ } & (\mathbf{YR})_{i,:}   \label{equ:approx-u}\\
    \mathbf{W} \approx \mbox{\ } & (\mathbf{S}_{:,j}) (\mathbf{Q}_{:,j})^T  \label{equ:approx-w}
\end{align}
where $\mathbf{P}$, $\mathbf{Q}$, $\mathbf{R}$, and $\mathbf{S}$ are results from the matrix factorization in Equations~\ref{equ:pq}--\ref{equ:z}. 

\subsection{Realtime recommendation by HI}
A realtime recommender responds to users' on-the-fly activities with new recommendations in realtime. A typical example of this application scenario is Next-Up video recommendations. Based on which items the user has interacted with in the immediately previous session, a realtime recommender comes up with new recommendations within milliseconds in reaction to the triggering events (e.g., clicks, long/short views, or likes). 
Being an extension of CF, HI algorithms' succinctness and flexibility make them an ideal candidate for realtime personalized recommendation engine. An example algorithm is outlined as follows.
\par
\mbox{Step 0. } \mbox{input matrix concatenating p2p and n2p parts:}
\begin{equation}
\begin{aligned}
\mathbf{H} = \begin{bmatrix} \mathbf{X}^T \mathbf{X} \\ \mathbf{Y}^T \mathbf{X}  \end{bmatrix}, \mbox{\ where\ } \mathbf{H} \in \Re^{2n\times n} \label{equ:h}
\end{aligned}
\end{equation}

\mbox{Step 1. } \mbox{rank-$k$ matrix factorization of $\mathbf{H}$: \ \ \ }
\begin{equation}
\begin{aligned}
\mathbf{H} \approx \mbox{\ } \mathbf{A} \mbox{\ } \mathbf{B}, \mbox{\ where\ } \mathbf{A} \in \Re^{2n\times k}, \mathbf{B} \in \Re^{k\times n} \label{equ:ab}
\end{aligned}
\end{equation}

\mbox{Step 2. } \mbox{smooth approximation of $\mathbf{H}$: \ \ \ }
\begin{equation}
\begin{aligned}
\mathbf{H}' = \mbox{\ } \mathbf{A} \mbox{\ } \mathbf{B},  \mbox{\ where\ } \mathbf{H}' \in \Re^{2n\times n}  \label{equ:h'}
\end{aligned}
\end{equation}

\mbox{Step 3. } \mbox{inferred p2p-\&-n2p recommendations per user: }
\begin{equation}
\begin{aligned}
^*\mathbf{G}_{i,:} = \mbox{\ } [\mathbf{X}, \beta \mathbf{Y}]_{i,:} \mbox{\ } \mathbf{H}', \mbox{\ where\ } ^*\mathbf{G} \in \Re^{m\times n} \label{equ:g}
\end{aligned}
\end{equation}
Step 0--2 correspond to offline model training. $\mathbf{H}$ contains both p2p and n2p inference components, therefore the MF on $\mathbf{H}$ is able to model p2p and n2p as well as the interaction between them. Step 3 performs online (realtime) inference for new recommendations for current user $i$. 

\section{Experiments}
\label{sec:experiment}

\begin{table*}[ht]
\footnotesize
  \centering
    \begin{tabular}{| r | r | r | r || r | r | r | r || r | r | r |}
		\hline
		\multicolumn{4}{|c||}{AUC-ROC, all users} & \multicolumn{4}{c||}{AUC-ROC, D1--D7 users} & \multicolumn{3}{c|}{mAP} \\
		\hline
	    CF-NN & CF-MF & CF-DA & \textbf{HI-NN} & CF-NN & CF-MF & CF-DA & \textbf{HI-NN} & CF-NN & CF-MF  & \textbf{HI-NN} \\
		\hline
	$0.616$  & $0.589$  & $0.577$  & \textbf{0.765} & 0.561  & $0.557$  & $0.550$ & \textbf{0.734} & $0.525$  & $0.501$  & \textbf{0.662} \\ 
		\hline
	\end{tabular}
    \label{tab:auc-all}
\caption{Test-BT. (Left tab): AUC, all users. (Middle tab): AUC, D1--D7 users. (Right tab): mAP. } 
\label{tab:testbt}
\end{table*}

This section evaluates and compares our approach, HI, with other state-of-the-art recommendation algorithms with and without diversity enhancement techniques on a well-known public datasets (MovieLens dataset) and a down-sampled real-world application production data. 

\subsection{Evaluation metrics}
Recommendation relevance (traditional recommendation quality) is measured by AUC ROC (area under the ROC curve), mAP (mean average precision) of the model against testing data, precision/recall, and empirical engagement metric in actual production environment (in Test-BT).
\par
By a widely-adopted definition of recommendation diversity, the diversity between two items $i$ and $j$ is calculated as $Diversity_{(i,j)} = 1-Similarity_{(i,j)}$~\cite{bradley2001improving,castagnos2013diversity,l2014understanding}. $Similarity_{(i,j)}$ is obtained as the inferred inter-item similarity $^*x_{i,j}$ in Equation~\ref{equ:x} from conventional CF. Then, the diversity of each user, i.e., the diversity of the top-$n$ items (by inferred ratings) recommended to the user, is computed as the average diversity over all pairs on the recommendation list for the user, i.e., $Diversity_{\mbox{user}} = \sum_{i,j \in \mbox{recommendations2user}}Diversity_{(i,j)}/(2n(n-1))$.

\subsection{Algorithms and tests}
\textit{\mbox{\ \ \ \ }HI algorithms in the tests are as follows}. 

\textbf{HI-RT}: realtime personalized recommendation by HI's MF implementation, as outlined in Equations~\ref{equ:h}--\ref{equ:g}. 

\textbf{HI-NN}: candidate generation by HI's candidate generation, as outlined in Equations~\ref{equ:rs}--\ref{equ:z}; ranking by deep neural network, as outlined in Figure~\ref{fig:hi-architecture} and Equation~\ref{equ:lossfunc}. HI candidate generation assumes a mixture ratio of p2p:n2p = 67:33.

\textit{Conventional CF algorithms in the tests are listed below}. 

\textbf{CF-RT}: realtime personalized recommendation by p2p MF, as outlined in Equations~\ref{equ:pq}--\ref{equ:x}.

\textbf{CF-MF}: candidate generation and ranking by p2p MF, as outlined in Equations~\ref{equ:pq}--\ref{equ:x}.
\par
\textbf{CF-NN}: candidate generation by p2p MF; ranking by deep neural network, as outlined in Figure~\ref{fig:hi-architecture} and Equation~\ref{equ:lossfunc} with only the p2p-related components: $^+f_U$, $^+f_V$, $^+\mathcal{L}$.
\par
\textbf{CF-DA}: candidate generation and ranking by p2p MF; followed by diversity-aware re-ranking, where the re-ranking score is calculated as a weighted sum of relevance score and diversity score: $Relevance + \phi Diversity$. This represents a large group of ranking score augmentation algorithm for diversity improvement~\cite{bradley2001improving,castagnos2013diversity,l2014understanding}. 
\par
\textbf{CF-DM}: candidate generation and ranking by p2p MF; followed by diversity-maximizing re-ranking, where the re-ranking selects $n$ final recommendations that maximize the overall diversity (from the top-$n'$ items by p2p MF), where $n'=5n$. This represents a group of diversity maximizing algorithms~\cite{divers11Adomavicius,acm16paudel}. In our experiments, $n=10$ for all algorithms, $n'=50$ for CF-DM.
\par
Our choice of MF implementation is Apache Spark ALS, for its proven effectiveness and multi-thread capability. The target rank of MF in all algorithms is set to $k=10$.    
\par
In the two tests, we compare algorithms in their pure CF form, without using any content-based features in the models. This is for keeping the comparison on a clean common ground and for keeping the focus on the essence of approaches. This leaves a large space for improvement beyond the models in the test, by enriching the model features and/or including algorithms beyond CF.  
\par
\textbf{Test-RT:} Use-case: realtime personalized recommendation; Data: MovieLens dataset. The whole dataset is divided into a training dataset and a test dataset (on a 80:20 training:test ratio), based on the timestamp of every event/example. Overall, MovieLens contains over $6000$ user and around $4000$ items.  
\par
\textbf{Test-BT:} Use-case: batch personalized recommendation, including candidate generation and ranking; Data: a down-sample of production data from the recommender on the Explore-like page of a real-world user-generated multimedia content sharing platform. The dataset is obtained by down-sampling production data during a couple of weeks' time, which contains around one million unique users and one million unique video items. As in Test-RT, examples in the training set are strictly older than those in the test set in time. In this use case, the feedback from a user to an item indicates whether the user long-watches the video (watch more than a certain percentage of the video).

\subsection{Results and observations}
Experiment results are organized under the two tests. 
\par
\textbf{Test-RT}. There are three algorithms in this test: CF-RT, CF-DM, and HI-RT. Overall, HI-RT outperforms both baseline algorithms: CF-RT and CF-DM simultanuously on all performance metrics: AUC ROC, precision (at a given recall value), recall (at a given precision value), diversity median, and diversity 25 percentile.

\begin{table}[t]
\footnotesize
\centering
\begin{tabular}{| l | r | r | r |}
	\hline
	          & CF-RT & CF-DM & \textbf{HI-RT} \\
	\hline
	AUC-ROC   & $0.556$ & $0.556$ & \textbf{0.669}  \\  
	\hline
	Precision (recall=$0.8$)  & $0.847$ & $0.847$ & \textbf{0.875}   \\ 
	\hline
	Recall (precision=$0.85$) & $0.78$ & $0.78$ & \textbf{0.91}   \\ 
	\hline
	Diversity median   & $0.294$ & $0.554$  & \textbf{0.751}   \\ 
	\hline		
	Diversity 25\%ile  & $0.269$ & $0.523$  & \textbf{0.723}   \\ 
	\hline
\end{tabular}
  \caption{Test-RT. Relevance and diversity results}
  \label{tab:movielensperf}
\end{table}

\begin{table}[t]
    \footnotesize
    \centering
	\begin{tabular}{| l | r | r | r |}
		\hline
		                    & CF-NN & CF-DA & \textbf{HI-NN} \\
		\hline
		Diversity median    & $0.226$ & $0.2352$ & \textbf{0.266}   \\ 
		\hline
		Diversity 25\%ile   & $0.1006$ & $0.1078$ & \textbf{0.1243}   \\ 
		\hline
	\end{tabular}
	  \caption{Test-BT. Diversity results}
      \label{tab:diversity}
\end{table}

Table~\ref{tab:movielensperf} summarizes the performance metrics from all three algorithms. HI has significant advantage over the baselines on diversity as well. CF-DM reduces the diversity advantage of HI-RT from $155\%$ down to $36\%$, at the cost of lower precision and recall in the final recommendations (which is not factored into our performance measurement, allowing an extra benefit for CF-DM baseline algorithm).
HI has $20\%$ higher AUC, $16.7\%$ higher recall, and $3.3\%$ higher precision than the baselines. The large advantage on recall shows HI has higher coverage due to divergent relevance.    
\par
\begin{figure}[t]
\centering
\begin{subfigure}{.29\textwidth}
  \centering
  \includegraphics[width=.95\linewidth]{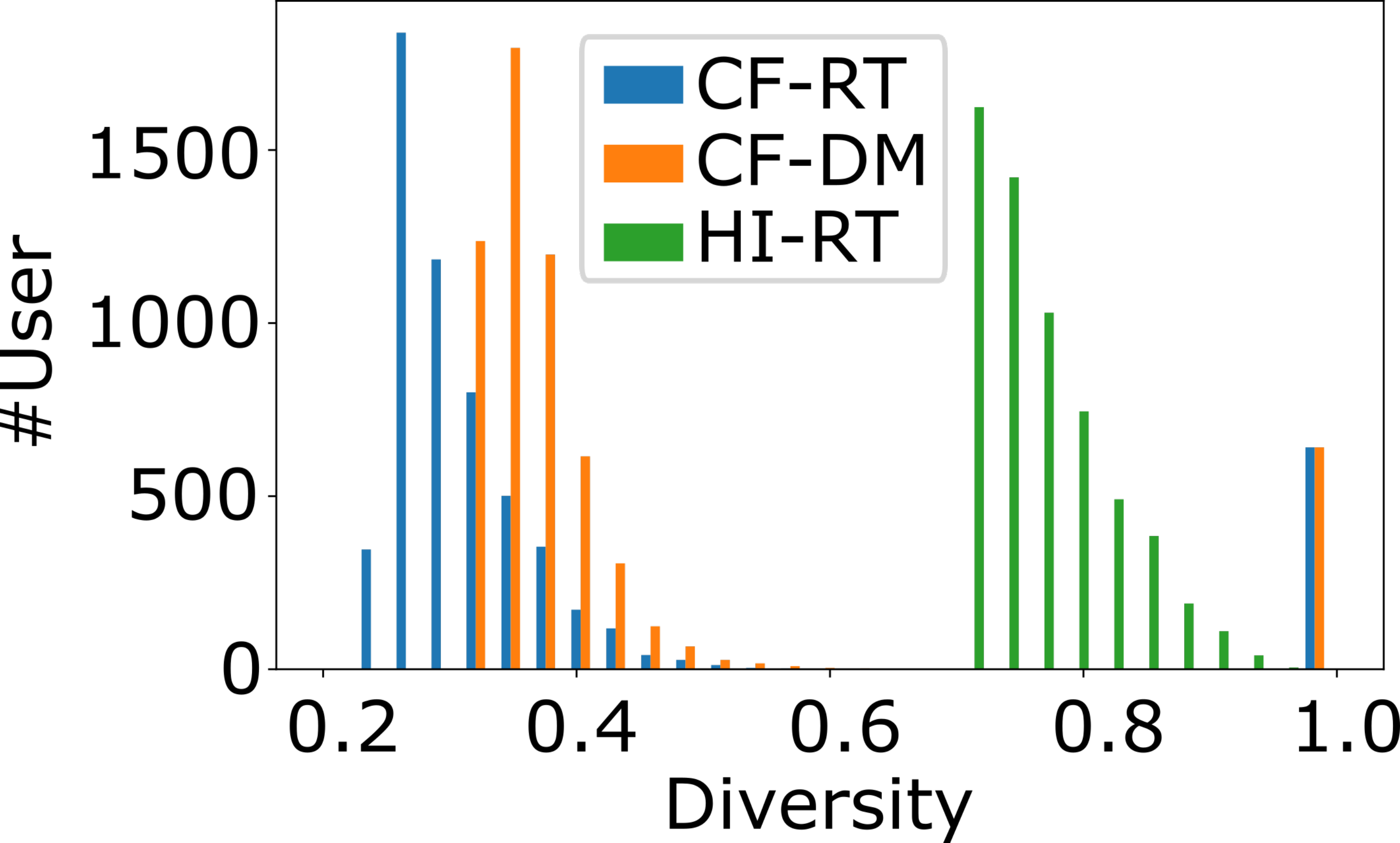} 
\end{subfigure}
\hfill
\begin{subfigure}{.16\textwidth}
  \centering
  \includegraphics[width=.95\linewidth]{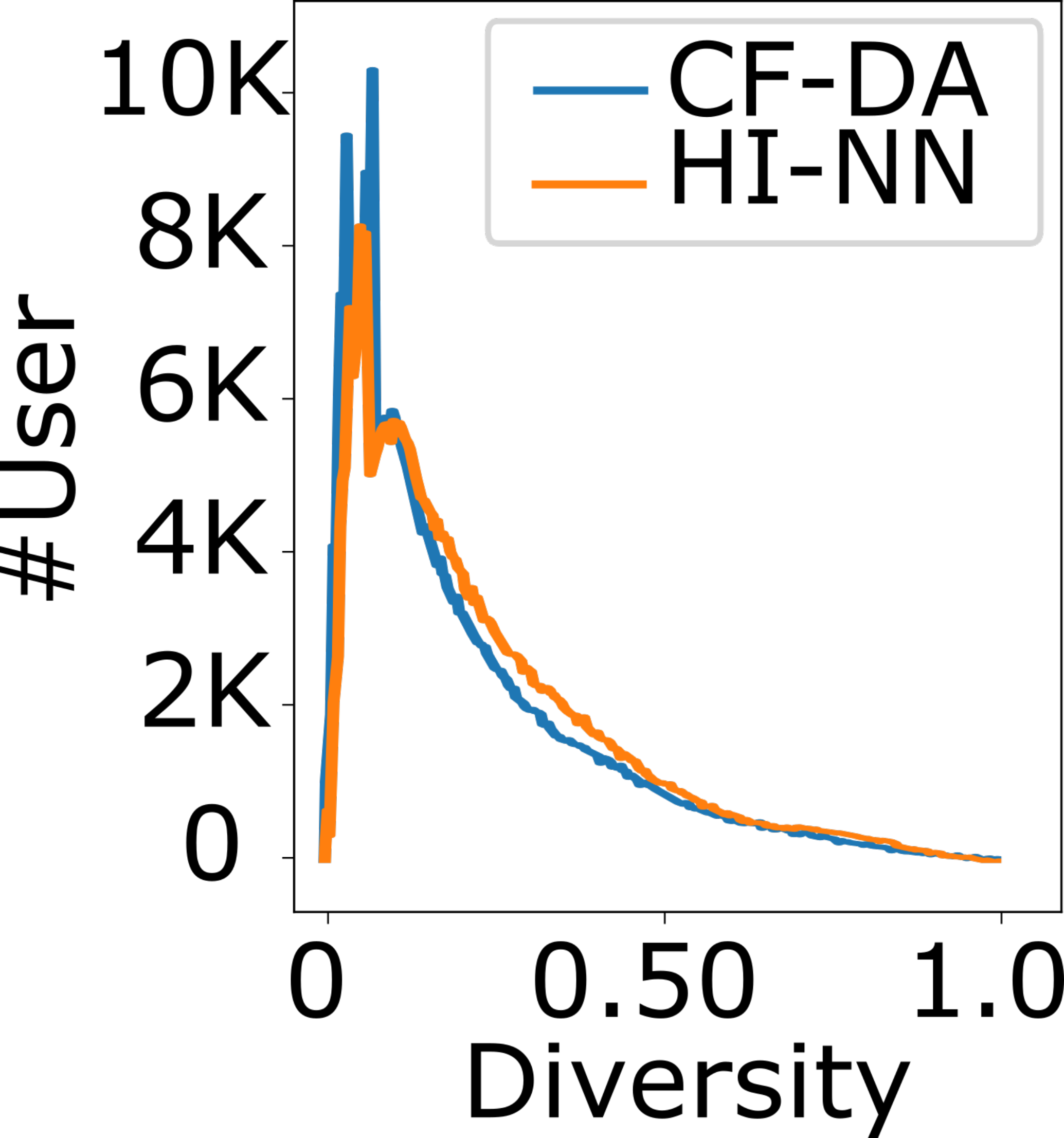}
\end{subfigure}
\caption{User-level diversity distribution over all users. \\ (Left figure): Test-RT.  (Right figure): Test-BT.}
\label{fig:diversity-result}
\end{figure}

Detailed diversity distribution over users can be found in the left chart of Figure~\ref{fig:diversity-result}. The user-level recommendation diversity in CF-RT and CF-DM concentrates around $0.3$--$0.4$, with spikes around $1.0$, which correspond to users having little positive feedback in the past. Conventional CF without n2p inference cannot come up with meaningful recommendations for those users, i.e., recommendation for them are random selections, thus the high diversity. 
\par
Recommendation processing latency: HI-RT and CF-RT take $4ms$ on average; CF-DM takes around $7ms$ (due to its extra re-ranking processing), in a multi-thread program. 
\par
\textbf{Test-BT}. There are four algorithms in this test: CF-MF (representing the conventional CF by MF algorithms), CF-NN (conventional CF enhanced by deep neural network), CF-DA (conventional CF with diversity-aware re-ranking), and HI-NN (HI candidate generation and ranking). 

Overall, HI-NN outperforms all baseline algorithms (CF-NN, CF-MF, and CF-DA) in both relevance and diversity, across model performance evaluation and online A/B tests. 
The AUC results are shown in the left and middle tabs of Table~\ref{tab:testbt} for all users and (day 1--7) new users, respectively. Note that HI's performance advantage further expands when the audience changes from all users to new users. This implies that HI is also helpful for user cold start. Similar improvement on mAP can be seen in the right tab of Table~\ref{tab:testbt}. 
\par
Table~\ref{tab:diversity} shows diversity measurement of the algorithms. With the addition of diversity metric in the re-ranking step, CF-DA does improve recommendation diversity over other CF algorithms, but still has lower diversity than HI-NN, at a much lower AUC than HI-NN.  
The right chart in Figure~\ref{fig:diversity-result} shows the user-level recommendation diversity distribution over users: CF-DA vs. HI-NN. 
\par
In a two-way A/B test on the real-world production, HI-NN achieved a \textbf{$32.05\%$} lift over CF-NN on a production engagement performance metric, long-watch. 
\par
All algorithms run in a cluster of 15 nodes, each of which is equipped with 20 CPU cores and 250 GB memory. The model training runtimes and inference runtimes are similar across the algorithms in this test. 


\section{Conclusion}
\label{sec:conclusion}

We presented heterogeneous inference (HI) to achieve divergent relevance, where diversity and relevance are two inherent outcomes of one negative-to-positive driven inference process. Meanwhile, HI's generality and succinctness allows it to be applied to various recommendation scenarios/use-cases, including realtime personalized recommender. 

\bibliography{main_arxiv}

\end{document}